%% file: main.tex
\documentclass[letterpaper, 10 pt, conference]{ieeeconf}  

\IEEEoverridecommandlockouts                              

\usepackage{cite}
\usepackage{balance}
\usepackage{multicol}
\usepackage{subcaption}
\usepackage{comment}
\usepackage{url}
\usepackage{graphicx}
\graphicspath{{../Figures/}}
\usepackage{amsmath}
\usepackage{amssymb}

\usepackage{amsthm}
\usepackage{nameref}
\usepackage{color}
\usepackage{comment}
\usepackage{mathtools}

\input wenmacro.tex


\title{Adaptive Neural Trajectory Tracking Control for Flexible-Joint \\ Robots with Online Learning 
}

\author{Shuyang Chen$^{1}$, 
John T. Wen$^{2}$
\thanks{$^{1}$Shuyang Chen is with the Department of Mechanical, Aerospace, \& Nuclear Engineering, Rensselaer Polytechnic Institute, 110 8th St, Troy, NY 12180 USA
        {\tt\small chens26@rpi.edu}}%
\thanks{$^{2}$John T. Wen is with the Department of Electrical, Computer \& Systems Engineering, Rensselaer Polytechnic Institute, 110 8th St, Troy, NY 12180 USA
        {\tt\small wenj@rpi.edu}}%
}

\begin{document}

\maketitle
\thispagestyle{empty}
\pagestyle{empty}

\begin{abstract}

Collaborative robots and space manipulators contain significant joint flexibility.  It complicates the control design, compromises the control bandwidth, and limits the tracking accuracy. 
The imprecise knowledge of the flexible joint dynamics compounds the challenge.
In this paper, we present a new control architecture for controlling flexible-joint robots.  
Our approach uses a multi-layer neural network to approximate unknown dynamics needed for the feedforward control.  
The network may be viewed as a linear-in-parameter representation of the robot dynamics, with the nonlinear basis of the robot dynamics connected to the linear output layer. 
The output layer weights are updated based on the tracking error and the nonlinear basis. 
The internal weights of the nonlinear basis are updated by online backpropagation to further reduce the tracking error.
To use time scale separation to reduce the coupling of the two steps -- the update of the internal weights is at a lower rate compared to the update of the output layer weights. 
With the update of the output layer weights, our controller adapts quickly to the unknown dynamics change and disturbances (such as attaching a load). The update of the internal weights would continue to improve the converge of the nonlinear basis functions.
We show the stability of the proposed scheme under the ``outer loop'' control, where the commanded joint position is considered as the control input.
Simulation and physical experiments are conducted to demonstrate 
the performance of the proposed controller on a Baxter robot, which exhibits significant joint flexibility due to the series-elastic joint actuators. 
\end{abstract}

\section{INTRODUCTION}
\label{introduction}
Robot manipulators are playing important roles in an increasing range of applications such as material handling, assembly, surface finishing, surgery, and in-space satellite servicing. There has been numerous work on the trajectory tracking control of
 robot manipulators~\cite{qu1995robust}. If the robot dynamics is expressed in the linear-in-parameter form, adaptive controller has been developed to achieve asymptotic tracking~\cite{slotine1987adaptive}. However, accurate dynamics models of robot manipulators are rarely available, 
 particularly for flexible-joint manipulators.  For such robots, a simplified reduced-order model may be used~\cite{Spong1987} 
 for feedforward control, with model parameters obtained from offline identification.
 For rigid robots, iterative learning control (ILC)~\cite{arimoto1990learning} has been applied for tracking specific trajectories,
 but the learned representation is not transferable to new trajectories. Neural network (NN) based controller demonstrates high generalization capability to new trajectories given enough training data and carefully designed architecture~\cite{Miyamoto1988}. These control approaches have been extended to flexible-joint robots
 \cite{brogliato1995global,wang1995simple,lee1998adaptive}, but the control performance is sacrificed to guarantee safety, especially for the high speed motion.

Numerous schemes have been proposed to control a robot with unknown dynamics, and examples include dynamics approximation by wavelet networks~\cite{Lin2006}, Gaussian Processes~\cite{Williams2009}, Locally Weighted Projection Regression (LWPR), fuzzy logic systems~\cite{HO2007801} and neural networks~\cite{Jiang2017}. NNs are used to either approximate the robot forward dynamics~\cite{Perez-Cruz2014} or inverse dynamics~\cite{Talebi1998} for the controller design. In our previous work~\cite{IROS2019}, we developed two feedforward controllers using recurrent neural networks (RNNs) to compensate for the unknown dynamics of a Baxter robot. One controller was based on a unidirectional RNN that approximates the forward dynamics of Baxter and the other controller was based on a bidirectional RNN that approximates the non-causal dynamical system inverse of Baxter. Though the approach has demonstrated good tracking results, there are several drawbacks. 
The RNNs require a large amount of training data to obtain a satisfactory approximation of the robot dynamics. The approach also lacks a rigorous trajectory error convergence proof and stability analysis. In addition, the NNs were trained offline with no online updates. Similarly, in~\cite{chen2019}, we trained a multi-layer feedforward NN to approximate the dynamical inverse of an ABB industrial robot with training data collected by implementing ILC with a large amount of trajectories. The trained NN performed well in compensating for the unseen trajectories but has the similar issues to~\cite{IROS2019}. In~\cite{Narendra1990}, feedforward NNs and RNNs were evaluated for identification and control of structured dynamical systems. The parameters of NNs were updated by online backpropagation and controllers were designed using the trained NNs. The parameters update of NNs and controller implementation were conducted simultaneously but at different rates. 
However, no stability proof was provided. A survey of neural-learning robot manipulators control can be referred to in~\cite{JIN201823}.

Adaptive control~\cite{slotine1987adaptive} has long been used to achieve globally asymptotically trajectory tracking and the approach is based on expressing robot dynamics in a linear-in-parameter form. Essentially, it linearizes the system through controller online adaptation to cancel the nonlinear dynamics. 
Neural network is a natural candidate to approximate the robot dynamics for adaptive controllers development~\cite{Yoo2008}.  A popular choice is the radial basis function (RBF) NN~\cite{Wang2017NeuralL, He2018NNLS} since it naturally expresses the approximation linearly in the parameters. In~\cite{Yang2017}, an adaptive controller was developed based on RBF NNs to compensate for the unknown dynamics and payload for a Baxter robot. The controller produced control inputs in the joint-torque level. However, for many industrial robots, users have only access to the joint position or velocity setpoint control. Further, it is non-trivial to select the centers and shapes of radial basis functions, and typically a large amount of Gaussian kernels are required to approximate the complex robot dynamics accurately.
Finally, some properties of robot dynamics cannot be guaranteed with individual RBF NN approximation, such as the positive definiteness of the inertia matrix and passivity. In~\cite{Ranatunga2015,Lewis1996}, an adaptive controller in the joint-torque level based on a two-layer feedforward perceptron NN was developed for control of robot manipulators with a guaranteed tracking performance. The dynamics approximation was nonlinear in the unknown parameters and Taylor series expansion was used for development of the parameters adaptation law. A robust control term was introduced to overcome the NN approximation errors and system uncertainties. With more NN layers, the adaptation law derivation would be far more complex. Other proposed control frameworks that use general function approximators to learn system dynamics for linearizing controllers design can be referred to in~\cite{2018arXiv181108027S, Noormohammadi2018,Umlauft2019}.


\begin{figure}[tb]
\centering
\includegraphics[width=\linewidth]{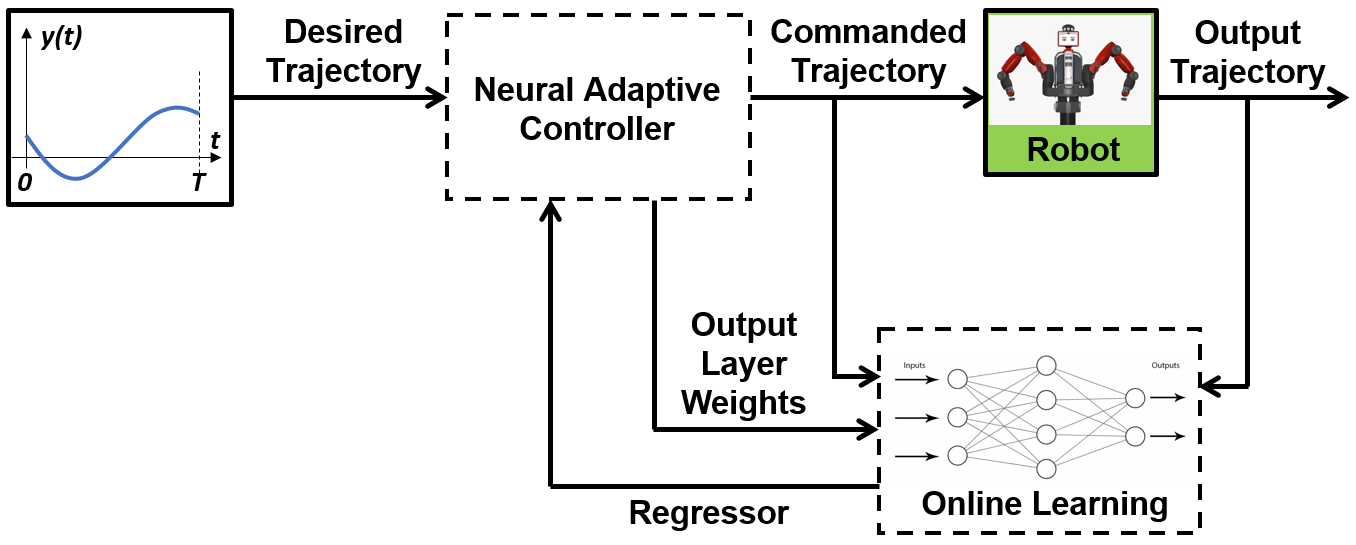}
\caption{Overall control architecture.}
\label{fig:overall_structure}
\end{figure}

This paper presents a novel neural adaptive controller for the tracking control of a flexible-joint robot under the outer-loop control.  
Fig.~\ref{fig:overall_structure} shows the schematics  of the control architecture. We design a network which is composed of a regressor network and an output layer. Thus the NN parameters can be classified into two categories: the internal weights of the regressor, and the output layer weights. The goal is to use the network to emulate the linear-in-parameter form of the robot dynamics. The updates of the internal and output layer weights are parallel and coupled, and operating at different time scales. The output layer weights are updated at a fast rate by the adaptation law developed from the Lyapunov analysis, and the internal weights are updated slowly via online backpropagation using collected robot input/output data. During the backpropagation, the output layer weights of the network remain fixed so that only the regressor weights are updated. The updated regressor is then involved in the output layer weights adaptation. This time scale separation of the internal and output layer parameters is inspired by the singular perturbation theory. 




The rest of the paper is organized as follows. Section~\ref{problem statement} states the problem, followed by the description of controller design in Section~\ref{methodology}. The simulation and physical robot experimental results are presented in Sections~\ref{simulation} and~\ref{experiment}, respectively. Finally, Section~\ref{conclusions and future work} concludes the paper.

\section{Problem Statement}
\label{problem statement}
We consider the trajectory tracking control problem for a flexible-joint robot manipulator modeled as~\cite{Wang2017NeuralL, He2018NNLS}:
\begin{equation}
\begin{split}
     M(\theta)\ddot{\theta} + C(\theta, \dot{\theta})\dot{\theta}+G(\theta)+K\cdot (\theta-\theta_m) = 0 \\
    J_m{\ddot{\theta}}_m+K\cdot (\theta_m-\theta) = \tau ,
    \label{eqn:original dynamics}
\end{split}
\end{equation}
where $\theta, \theta_m$ are the link position and motor position. $M(\theta), C(\theta, \dot{\theta})\dot{\theta}, G(\theta)$ denote the manipulator inertia matrix ($M$ is symmetric positive definite and $\dot{M}-2C$ is skew-symmetric), Coriolis and centrifugal torques, and gravitational torques, respectively. $\tau$ is the motor torque input, and $J_m$ and $K$ are constant diagonal and positive definite matrices representing the motor inertia and joint stiffness. 
To simplify the analysis, we make the following assumptions:
\begin{itemize}
    \item The motor position $\theta_m$ is perfectly controlled, i.e., we can ignore the motor dynamics and use $\theta_m$ as an input to drive $\theta$.
    \item Joint stiffness is the same for all joints, i.e., $K=k_p I$, $k_p>0$.
\end{itemize}
The robot dynamics can be rewritten as
\begin{equation}
    M(\theta)\ddot{\theta} + C(\theta, \dot{\theta}) \dot{\theta}+H(\theta)=k_p \theta_m ,
\label{eqn:dynamics}
\end{equation}
where $H(\theta) = G(\theta) + k_p \theta$. 
We then express the dynamics into a linear-in-parameter form as follows:
\begin{equation}
    Y(\dot\theta_1,\theta,\dot\theta_2,\ddot\theta) a  := k_p\inv (M(\theta)\ddot \theta + C(\theta,\dot\theta_2)\dot \theta_1 + H(\theta)) , 
    \label{linear form}
\end{equation}
where $Y$ is the regressor and $a$ is a constant vector.  
Our goal is to design an adaptive controller for $\theta_m$ that achieves globally asymptotically tracking of a desired trajectory $\theta_d$ with bounded first and second-order time derivatives. We will first test the approach in simulation with a single pendulum,
and then extend it to a multi-link robot such as the Baxter
and compare the performance of the adaptive controller with a baseline proportional-derivative (PD) controller (with gains $K_1$ and $K_2$) as below:
\begin{equation}
    \theta_m = \theta_d -K_1 (\theta-\theta_d)-K_2 (\dot{\theta}-\dot{\theta}_d).
\label{eqn:baseline PD}
\end{equation}
Finally, we compare the adaptive controller with and without the online regressor backpropagation to demonstrate the feasibility of the proposed regressor online learning scheme.

\section{Neural Adaptive Controller}
\label{methodology}
In this section, we propose a network architecture and design an adaptive controller that achieves asymptotically trajectory tracking for the robot manipulators modeled in~\eqref{eqn:dynamics}.
\subsection{Controller Design}
For the controller derivation, we use the following  Lyapunov-like function candidate for stability 
analysis. The design of NN architecture naturally arises during the stability analysis.
\begin{equation}
V = \frac{1}{2}s^T M s+\frac{1}{2}\widetilde{a}^T k_p P^{-1}\widetilde{a} ,
\label{eqn: derivation 1}
\end{equation}
where $M$ and $k_p$ are the robot inertia matrix and joint stiffness in~\eqref{eqn:dynamics} and $s$ is a sliding vector where $s = \dot{e}(t)+\Lambda e(t)$
with $e(t) = \theta(t) - \theta_d(t)$. $\widetilde{a} = \hat{a}-a$ and $\hat a$ is an estimate of the constant vector $a$. The designed weighting matrix $P$ controls the parameter adaptation rate and is symmetric and positive definite.
The Hurwitz matrix $-\Lambda$ makes $e(t), \dot{e}(t)$ converge to zero exponentially as $s$ converges to zero. 
$\theta_r$ is defined so that $ s = \dot \theta - \dot \theta_r$ (i.e., $\dot\theta_r = \dot \theta_d - \Lambda e)$. 

By taking the time derivative of $V$ and from~\eqref{eqn:dynamics} and~\eqref{linear form}, we have:
\begin{align}
\begin{split}
&\dot{V} = s^TM\dot{s}+\frac{1}{2}s^T\dot{M}s+\dot{\hat{a}}^Tk_pP^{-1}\tilde{a} \\
&    =s^T(M\ddot{\theta}-M\ddot{\theta}_r)+\frac{1}{2}s^T\dot{M}s+\dot{\hat{a}}^Tk_pP^{-1}\tilde{a} \\
&    = s^T(k_p\theta_m-H-C\dot{\theta}-M\ddot{\theta}_r) + \frac{1}{2}s^T\dot{M}s
    +\dot{\hat{a}}^Tk_pP^{-1}\tilde{a} \\
&    =s^T(k_p\theta_m-H-C\dot{\theta}_r-M\ddot{\theta}_r)+\frac{1}{2}s^T(\dot{M}-2C)s \\
&\qquad\qquad    +\dot{\hat{a}}^Tk_pP^{-1}\tilde{a} \\
&   = s^T (k_p\theta_m-H-C\dot{\theta}_r-M\ddot{\theta}_r) + \dot{\hat{a}}^Tk_pP^{-1}\tilde{a} \\
& = s\tr k_p (\theta_m  - Y(\dot \theta_r,\theta,\dot \theta,\ddot\theta_r)a) + 
\dot{\hat{a}}^Tk_pP^{-1}\tilde{a}.
\label{eqn: derivation 2}
\end{split}
\end{align}
If $Y$ is known, then we may design $\theta_m$ to cancel $Y a$ using the estimate $\hat a$ as
\def\sgn{{\mbox{\bf sgn}}}
\begin{equation}
    \theta_m = Y(\dot\theta_r,\theta,\dot\theta,\ddot\theta_r)\hat a - K_s s - k \sgn(s).
    \label{eq:thetam}
\end{equation}
The $K_s s$ term with $K_s>0$ ensures the tracking error convergence. The additional high gain feedback term $k \sgn(s)$ provides robustness with respect to the network modeling errors and noises.
It follows 
\begin{equation}
    \dot V = - s\tr k_p K_s s - kk_p\norm{s}_1 + k_p s\tr Y \tilde a + k_p {\dot{\hat a}}\tr P\inv \tilde a ,
\end{equation}
where $\norm{\cdot}_1$ denotes the vector 1-norm. By choosing the adaptation law of $\hat a$ as
\begin{equation}
\dot {\hat a}  = -P Y\tr(\dot\theta_r,\theta,\dot\theta,\ddot\theta_r) s,
\label{eq:adaptation}
\end{equation}
we have 
\begin{equation}
    \dot V = - s\tr k_p K_s s - kk_p\norm{s}_1.
\end{equation}
This implies $s\to 0$ asymptotically which in turn implies $e\to 0$ asymptotically.
In the next section, we design a neural network architecture to approximate $Y(\dot\theta_r,\theta,\dot\theta,\ddot\theta_r) a$. 

\subsection{Neural Network Design}

\subsubsection{Network Architecture}

\begin{figure}[tb]
\centering
\includegraphics[width=\linewidth]{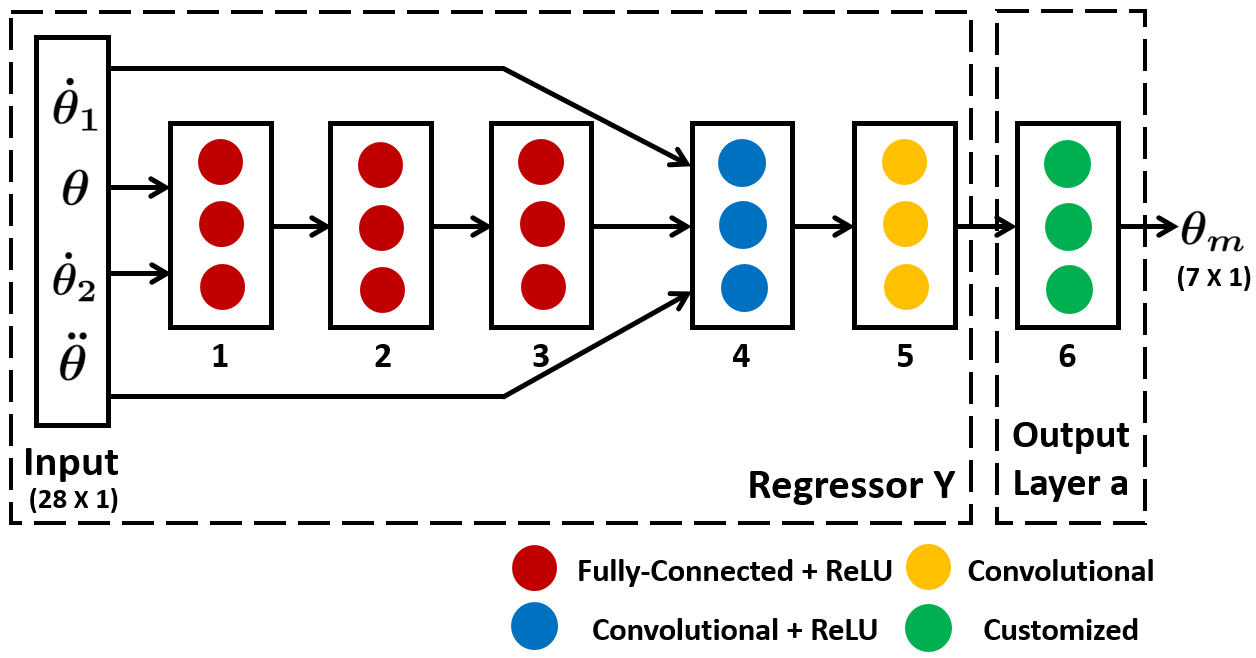}
\caption{Network architecture.}
\label{fig:network_architecture}
\end{figure}

\begin{table}[tb]
\caption{Neural network connections.}
\label{neuralnetwork analysis}
\begin{center}
\begin{tabular}{ccc}
\hline
\hline
Layer Number & Learnable Size & Output Size \\ 
\hline
1 & Weights: $100\times$14 & 100$\times$1 \\
 & Bias: $100\times$1 & \\
2 & Weights: $100\times$100 & 100$\times$1 \\  
 & Bias: $100\times$1 & \\
3 & Weights: $7\times$100 & 7$\times$1 \\
 & Bias: $7\times$1 & \\
4 & Weights: $1\times$1$\times$3$\times$100 & 7$\times$1$\times$100 \\
 & Bias: 1$\times$1$\times$100 & \\
5 & Weights: $1\times$1$\times$100$\times$200 & 7$\times$1$\times$200 \\
 & Bias: 1$\times$1$\times$200 & \\
6 & Weights: $200\times$1 & 7$\times$1 \\
\hline
\hline
\end{tabular}
\end{center}
\end{table}

We design a network as a directed acrylic graph composed of different layers, as presented in Fig.~\ref{fig:network_architecture}. Table.~\ref{neuralnetwork analysis} summarizes the connections between consecutive layers.
The regressor network $Y(\dot{\theta}_1, \theta, \dot{\theta}_2, \ddot{\theta})$ has its own internal weights from the fully-connected layers and convolutional layers, and $a$ is an $N-$parameter output layer which is compatible with the output dimension of $Y$. 
Note that we need one convolutional layer (the yellow block in Fig.~\ref{fig:network_architecture}) instead of a fully-connected layer as the final layer of $Y$, as the size of output of $Y$ is 2-dimensional (7 by $N$) for a 7-joint robot, whereas a fully-connected layer only provides a 1-dimensional vector output. Thus, for a single-input and single-output (SISO) system such as a single pendulum, the last layer of $Y$ is a fully-connected layer.

\subsubsection{Training Data Collection}

We collect data from the left arm of the Baxter robot by commanding 30 sinusoidal trajectories $\theta_m$ with different magnitudes and frequencies. We choose the trajectories by obeying the Baxter joint limits and also try to cover a feasible portion of its joint workspace. Each trajectory contains 2500 joint position setpoints for all 7 joints (thus a 7 by 2500 matrix) and is commanded to Baxter at 100 Hz. The output joint position trajectory $\theta$ as well as the joint velocity trajectory $\dot{\theta}$ are collected at the same rate. The joint acceleration trajectory $\ddot{\theta}$ is approximated by cubic spline interpolation of the joint velocity trajectory.

\subsubsection{Network Training for Initialization}

During training, the input to the network is $(\dot{\theta}(t), \theta(t), \dot{\theta}(t), \ddot{\theta}(t))$, and the output of the network is the commanded position $\theta_m(t)$. We use 80~\% of the collected data for network training and 20~\% for testing. We use \texttt{AdamOptimizer} with an initial learning rate of $1\times 10^{-3}$ to tune the parameters of the network by minimizing the mean squared error (MSE) between the predicted output and the actual output over a randomly selected batch of 256 samples in each training iteration. We use $L_2$ regularization (penalizing large weights) to avoid overfitting. The training process converges after 5 epochs.
After the initial network training, we use the output of the neural network with input of $(\dot{\theta}_r, \theta, \dot{\theta}, \ddot{\theta}_r)$ as $Y \hat a$ to implement the controller in~\eqref{eq:thetam}. The output layer weights are then updated using~\eqref{eq:adaptation}. 
On a slower time scale, the regressor $Y$ is updated using online backpropagation with regularization. 
%
Note that the network provides an estimate of the complete $Y$ matrix, so we may use $Y\tr$ directly  in the update law~\eqref{eq:adaptation}.

\subsection{Online Backpropagation for Regressor Update}

\begin{figure}[tb]
\centering
\includegraphics[width=0.46\textwidth]{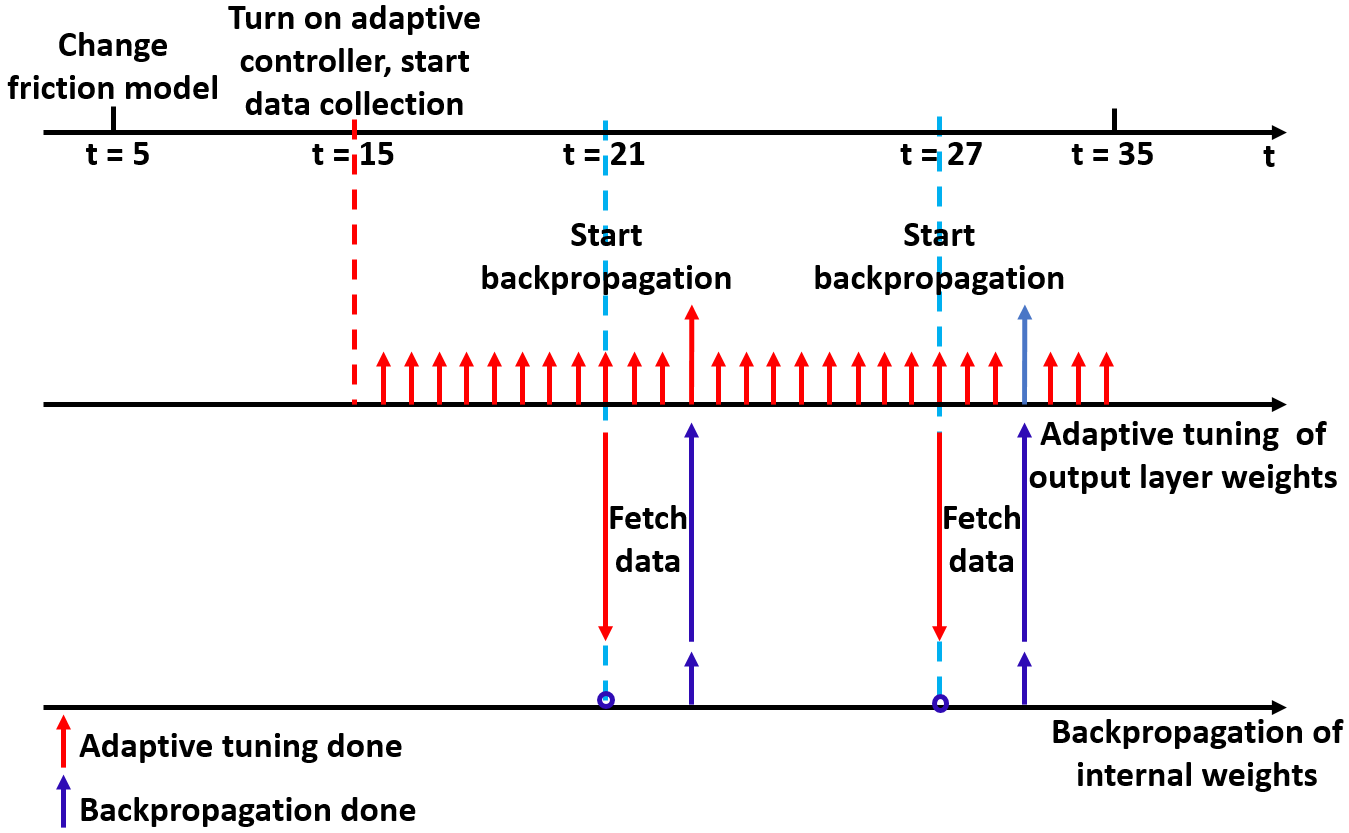}
\caption{Schematics of the regressor online learning.}
\label{fig:online_backpropa}
\end{figure}

When the system is subject to dynamics variation (e.g., attaching a load and friction change) and external disturbances, pure adaptation of output layer weights by~\eqref{eq:adaptation} may take a long time to make the tracking error converge. Here, we explore the online learning of regressor $Y$ with the collected input/output data of the system after dynamics variation. We update the output layer weights at a higher rate, whereas the regressor weights are updated at a lower rate inspired by the singular perturbation theory.
When updating the regressor, we fetch and freeze the values of the output layer weights and only update the regressor weights by backpropagation of the entire network. Then the updated regressor is utilized in~\eqref{eq:thetam} and~\eqref{eq:adaptation}. An example of using time scale separation for the regressor online learning is illustrated in Fig.~\ref{fig:online_backpropa}.

\section{Simulation}
\label{simulation}

To demonstrate the feasibility of the proposed controller with the regressor online learning scheme, we first simulate the scenario that the internal friction model changes from viscous and Coulomb friction to Stribeck friction of a single pendulum modeled in Simulink. The system dynamics is based on~\eqref{eqn:original dynamics} and the assumption that $K$ is a scalar naturally holds for this SISO system. Then, we test the approach with MIMO systems through physical experiments.

We collect 6000 training data sets ($\theta_m, \theta, \dot{\theta}, \ddot{\theta}$) from the single pendulum by commanding a Schroeder-phase multisine signal to the system. Then, a network with architecture denoted in Fig.~\ref{fig:network_architecture} is trained using MATLAB Deep Learning Toolbox with the following variations: first, the input size of the network is 4 by 1 and the output size is 1 by 1; second, the last layer of the regressor is a fully-connected layer instead of a convolutional layer, considering that the single pendulum is an SISO system. We simulate the following scenario for two systems with the same single pendulum model and the trained network, but in one system we implement regressor online learning (system~I), and as a comparison in system~II, no online learning is conducted. The same single pendulum in two systems is commanded to track a sinusoidal trajectory following the procedures as below:

\begin{itemize}
    \item From $t=0 \sim 5~s$, for both systems I and II, we implement the controller in~\eqref{eq:thetam} with the trained network but without the adaptation of $\hat{a}$ in~\eqref{eq:adaptation}. The value of $\hat{a}$ remains fixed and is from the initial value of the trained network.
    \item At $t = 5~s$, for both systems we change the friction model and with the same controller implemented as above.
    \item At $t = 15~s$, we turn on the adaptive controller (with adaptation law in~\eqref{eq:adaptation} activated) for both systems. We start collecting the input/output data of system I with the modified dynamics model.
    \item At $t = 21~s$, we start conducting the regressor online learning for system I using the collected input/output data every 6 seconds until end of the simulation. The first, second and third online updates complete at around $t = 21.5~s$, $t = 27.5~s$, and $t = 33.5~s$, respectively. Note that the controller in~\eqref{eq:thetam} keeps activated during the entire simulation for both systems.
\end{itemize}

\begin{figure}[tb]
\centering
\includegraphics[width=\linewidth]{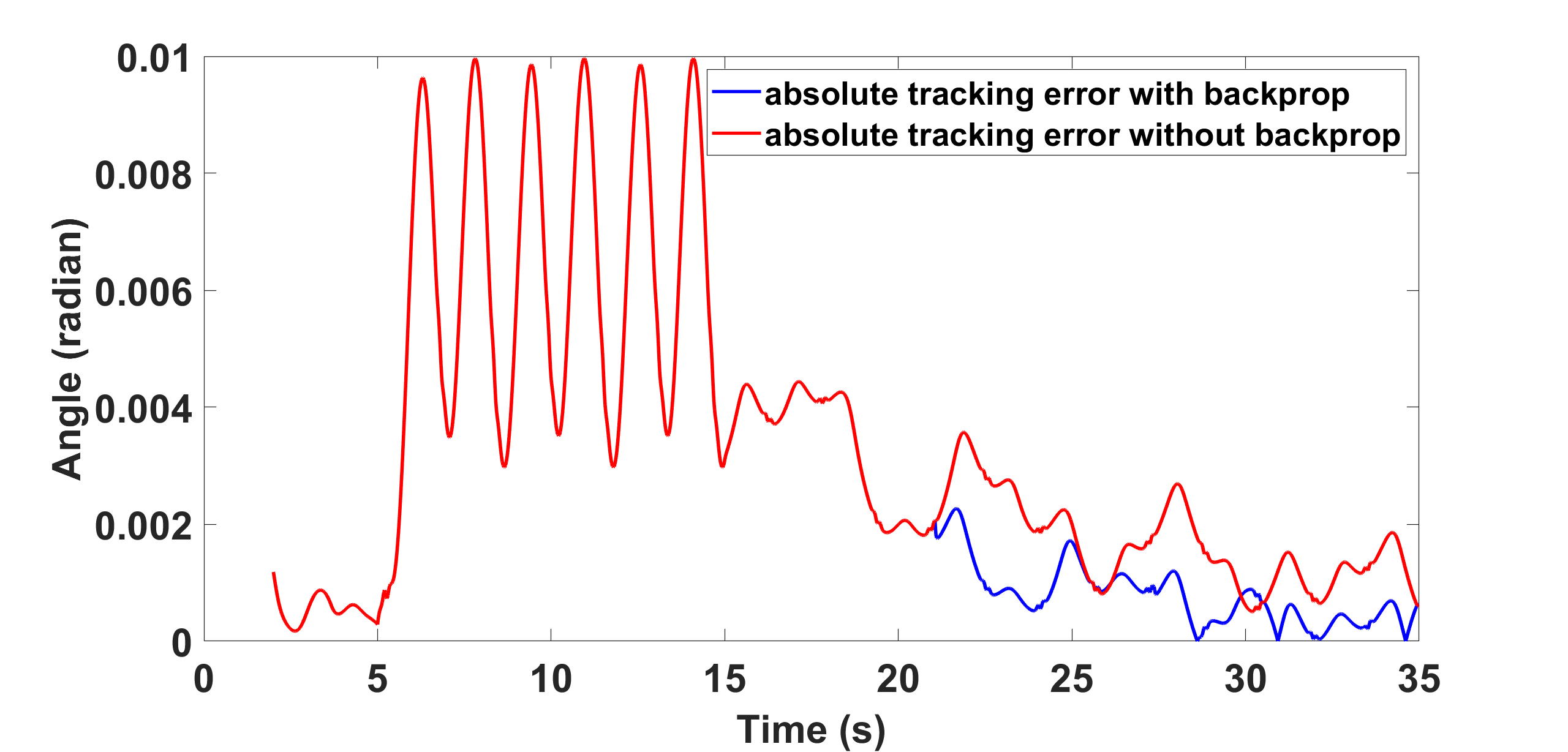}
\caption{Comparison of absolute tracking errors of a sinusoidal trajectory for the single pendulum with and without online backpropagation of the regressor. The large errors from $t = 0 \sim 2~s$ (the initial tracking error is 0.25 radian at $t = 0$ and the mean error is 0.0461 radian from $t = 0 \sim 2~s$) are not plotted to better visualize the later small errors.}
\label{fig:err_with_backprop}
\end{figure}

Figure~\ref{fig:online_backpropa} illustrates the timeline of system I in the simulation. The comparison of absolute tracking error in radian of two systems over the entire process is presented in Fig.~\ref{fig:err_with_backprop}. We do not plot the large initial tracking errors during $t = 0 \sim 2~s$ to better visualize the later smaller tracking errors.

The figure demonstrates the following stages:
\begin{itemize}
    \item The tracking errors converge for both systems from $t=0 \sim 5~s$, which means our trained network approximates the single pendulum dynamics model accurately.
    \item From $t=5 \sim 15~s$, the tracking errors of both systems increase due to the change of the friction model.
    \item Once we turn on the adaptive controller at $t=15~s$, the controller reduces the tracking errors for both systems. However, for system I, the tracking error is further reduced with the complete of regressor update via online backpropagation.
\end{itemize}

\section{Experiment with Baxter}
\label{experiment}

After an initial training of a network with architecture illustrated in Fig.~\ref{fig:network_architecture}, the testing of the control framework for MIMO systems such as a Baxter robot is composed of two parts. The first one is a feasibility study, in which we show the feasibility of our assumptions in Section~\ref{problem statement} by commanding the Baxter to track an unseen multi-joint sinusoidal trajectory and compare the tracking performance of the adaptive neural controller with the baseline PD controller in~\eqref{eqn:baseline PD}. In this study, there is no regressor online update implemented. The second part is a comparative study, in which we compare the performance of the adaptive controller with and without the regressor online learning by tracking a multi-joint sinusoidal trajectory with attaching a payload to the robot gripper. 

\subsection{Experiment I: Feasibility Study}
\begin{figure}[tb]
\centering
\includegraphics[width=\linewidth]{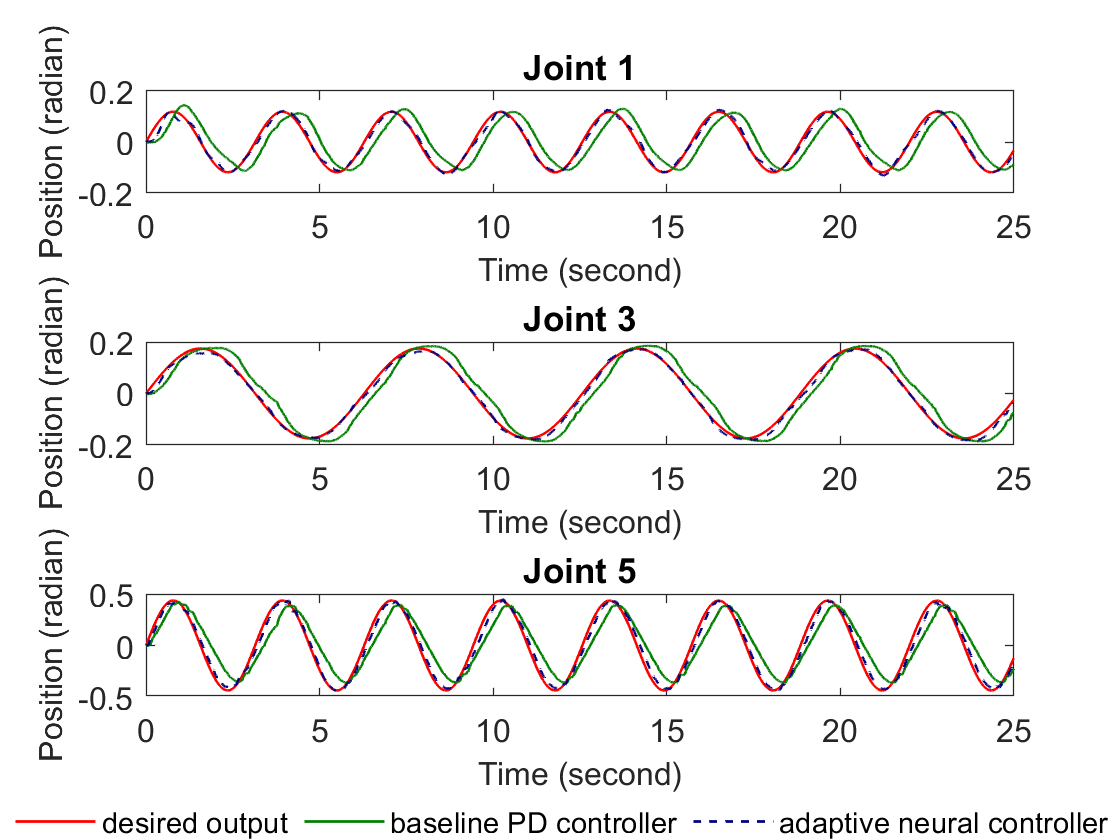}
\caption{Representative comparison of tracking performance with the baseline PD controller and the adaptive neural controller for sinusoidal joint position trajectories. See Table~\ref{error_7_joint_path} for a complete comparison for all 7 joints.}
\label{fig:track_multi_sin}
\end{figure}

Figure~\ref{fig:track_multi_sin} shows the representative results of joints 1, 3 and 5 for tracking an unseen sinusoidal joint position trajectory using the baseline PD controller in~\eqref{eqn:baseline PD} and the adaptive controller in~\eqref{eq:thetam} and~\eqref{eq:adaptation}. The controller parameters are chosen as
$$
\begin{aligned}
&(K_1,K_2) =(0.2,0.1), \,\, k=5,\,\, P = 0.05\\
&\Lambda=\mbox{diag}(3, 6, 3, 20, 10, 10, 10)\\
&K_s=\mbox{diag}(0.1, 0.01, 0.01, 0.1, 0.01, 0.01, 0.01).
\end{aligned}
$$

As a complete comparison, Table~\ref{error_7_joint_path} lists the corresponding tracking errors in terms of $\ell_2$ and $\ell_\infty$ norms (using the error vectors at each sampling instant) for all 7 joints. From the figure and the table, it is clear that the adaptive neural controller outperforms the PD controller (in average over 60~\% of improvement for all 7 joints).

\begin{table}[tb]
\caption{$\ell_2$ and $\ell_\infty$ norm of tracking errors of the sinusoidal joint trajectories in Fig.~\ref{fig:track_multi_sin} using the baseline PD controller and the neural adaptive controller.}
\label{error_7_joint_path}
\begin{center}
\begin{tabular}{ccc}
\hline
\hline
 & Baseline Controller & Adaptive Neural Controller \\
Joint & Tracking Error (rad) & Tracking Error (rad)  \\ 
 & $\ell_2 \hspace{1.5cm} \ell_\infty$ & $\ell_2 \hspace{1.5cm} \ell_\infty$ \\
\hline
1 & 2.7368 \hspace{0.5cm} 0.0885 & 0.5717 \hspace{0.5cm} 0.0473 \\
2 & 1.8055 \hspace{0.5cm} 0.0715 & 0.5204 \hspace{0.5cm} 0.0311  \\
3 & 2.0676 \hspace{0.5cm} 0.0706 & 0.5034 \hspace{0.5cm} 0.0275  \\
4 & 2.6028 \hspace{0.5cm} 0.0955 & 0.4335 \hspace{0.5cm} 0.0290  \\
5 & 7.4070 \hspace{0.5cm} 0.2534 & 2.3770 \hspace{0.5cm} 0.1064  \\
6 & 5.9475 \hspace{0.5cm} 0.2274 & 1.6294 \hspace{0.5cm} 0.0803  \\
7 & 4.7206 \hspace{0.5cm} 0.1873 & 2.2645 \hspace{0.5cm}  0.1278 \\
\hline
\hline
\end{tabular}
\end{center}
\end{table}

\subsection{Experiment II: Comparative Study}

We then conduct a comparative study to demonstrate the
effectiveness of the regressor online learning scheme with a $\sim$0.6~kg payload attached to the robot gripper (the maximum payload of the gripper is around 5 pounds). We compare the tracking results of two systems with (system I) and without (system II) the regressor online learning by following the procedures as below:

\begin{figure*}[tb]
\centering
\includegraphics[width=\linewidth]{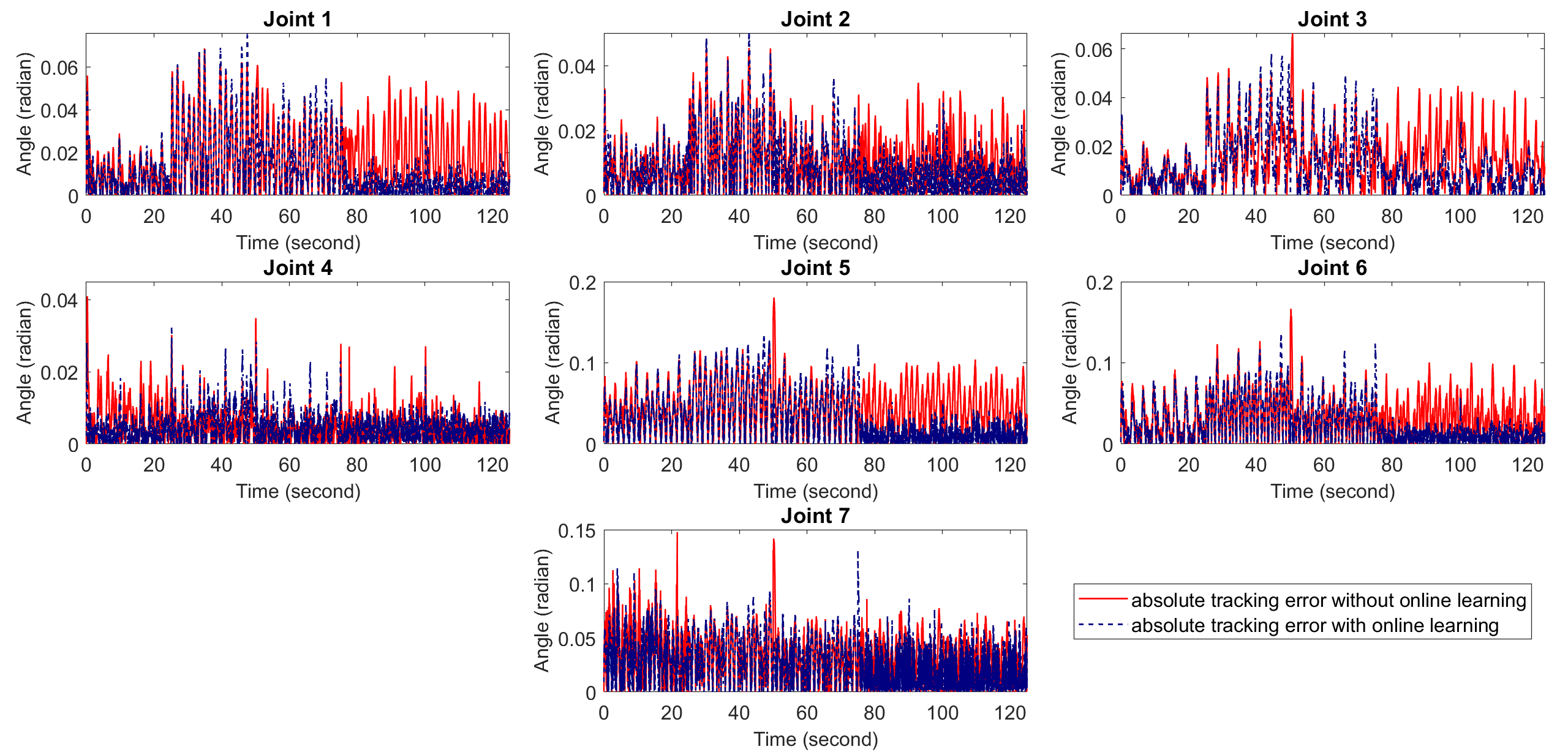}
\caption{Comparison of absolute tracking error using the developed adaptive controller with (blue curve) and without (red curve) the regressor online learning for tracking sinusoidal joint position trajectories.}
\label{fig:online_learning_7_joints_with_load} 
\end{figure*}

\begin{itemize}
    \item From $t=0 \sim 25~s$, for both systems I and II, we use the controller in~\eqref{eq:thetam} but without the adaptation of $\hat{a}$ in~\eqref{eq:adaptation}. The value of $\hat{a}$ remains fixed and is from the trained network.
    \item At $t = 25~s$, for both systems we attach the load to the robot gripper and implement the same controller as above.
    \item At $t = 50~s$, we turn on the adaptive controller (with adaptation law in~\eqref{eq:adaptation} activated) for both systems. We start collecting the input/output data of system I.
    \item For system I, we start implementing the regressor online update using the collected data every 25 seconds and the online updates are finished at $t = 75~s$ and $t = 100~s$, respectively. We find that the network weights converge after these two online updates.
\end{itemize}

Figure~\ref{fig:online_learning_7_joints_with_load} illustrates the comparison of the absolute tracking errors of two systems for the entire process. The error increases between $t = 25~s \sim 50~s$ after attaching the load. After turning on the adaptive controller at $t=50~s$, the tracking errors of both systems decrease but the system I with online learning (dashed blue curve in Fig.~\ref{fig:online_learning_7_joints_with_load}) reduces further than system II (red curve in Fig.~\ref{fig:online_learning_7_joints_with_load}), starting from the first complete of online learning at $t=75~s$. Table~\ref{table:online learning} compares the Frobenius norm of tracking errors of all 7 joints at different stages for two systems. The table shows that two systems have similar tracking performance from $t = 0 \sim 75~s$ with the same condition. However, starting from $t = 75~s$, system I presents a better tracking performance than system II due to the regressor online learning.

The comparison shows that the adaptive controller is able to compensate for the unknown load with the pure adaptation of the output layer weights. In addition, the online learning of regressor further reduces the tracking error with better model approximation.

\begin{table}[tb]
\caption{Frobenius norm of tracking errors (in radian) of all 7 joints of the sinusoidal joint position trajectories in Fig.~\ref{fig:online_learning_7_joints_with_load} using the neural adaptive controller with (system I) and without (system II) online learning of regressor.}
\label{table:online learning}
\begin{center}
\begin{tabular}{cccccc}
\hline
\hline
 & $[0,25]s$ & $[25,50]s$ & $[50,75]s$ & $[75,100]s$ & $[100,125]s$ \\
 \hline
I & 2.1449 & 3.745 & 2.830 & 1.577 & 1.199 \\
II & 2.376 & 3.494 & 3.045 & 2.748 & 2.689  \\
\hline
\hline
\end{tabular}
\end{center}
\end{table}

\subsection{Discussions}
It remains an open problem to choose the proper update rate ratio between the regressor online learning and output layer weights adaptation. In the comparative experiment, we implement the regressor online update every 25 seconds whereas the output layer weights adapt at a rate around 100~Hz (thus a ratio of 2500:1). 
Potentially, the regressor online learning may be accelerated with a GPU.

In addition, the dynamics of a rigid-joint robot has a similar form to~\eqref{eqn:dynamics}:
\begin{equation}
    M(\theta)\ddot{\theta} + C(\theta, \dot{\theta})\dot{\theta}+G(\theta) = \tau ,
\end{equation}
where $\tau$ is the joint torque vector. By assuming the inner-loop torque control as
\begin{equation}
    \tau = -K_p \cdot (\theta-\theta_c) - K_d \dot{\theta} +G(\theta),
    \label{controller}
\end{equation}
where $\theta_c$ is the commanded joint position. It follows
\begin{equation}
    M(\theta)\ddot{\theta} + (C(\theta, \dot{\theta})+K_d) \dot{\theta}+K_p \theta = K_p \theta_c .
\end{equation}
By identifying the controller gains in~\eqref{controller} through collected data, we can relax the assumption of a scalar $K_p$, but use the identified $K_p$ to develop a similar neural adaptive controller.

\section{Conclusions and Future Work}
\label{conclusions and future work}

In this work, we proposed a novel neural adaptive controller 
that demonstrates asymptotically trajectory tracking for a flexible-joint robot with unknown dynamics.
Initialized by a simple offline training, the proposed neural network which is composed of a regressor and an output layer, emulates the linear-in-parameter form of the robot dynamics. The regressor network updates online at a slower rate compared to the adaptation of the output layer weights. By simulation and experiments, the proposed control framework improves the trajectory tracking performance of the system. With the regressor online learning, the tracking performance of the adaptive controller is further improved.

Future work includes the exploration of choosing a proper update ratio between the regressor network and the output layer. It is also interesting to test the proposed control framework with rigid-joint robot manipulators as discussed above. Finally, we plan to compare the proposed approach with other neural adaptive controllers.

\section*{Acknowledgement}

This research was funded in part by the New York State Empire State Development Division of Science, Technology and Innovation (NYSTAR) under contract C170141.


\bibliographystyle{IEEEtran}
\bibliography{ral.bib}

\end{document}

%% file: wenmacro.tex
\def\beq{\begin{equation}}
\def\eeq{\end{equation}}
\def\beqa{\begin{eqnarray}}
\def\eeqa{\end{eqnarray}}
\def\ba{\begin{eqnarray*}}
\def\ea{\end{eqnarray*}}
\def\bc{\begin{center}}
\def\ec{\end{center}}
\def\ma{\left[ \begin{array}}
\def\ema{\end{array}\right]}

\def\tr{ ^T }

\def\inv{ ^{-1} }

\def\norm#1{\left\| {#1} \right\|}